\documentclass[runningheads]{llncs}

\usepackage[T1]{fontenc}
\def\doi#1{\href{https://doi.org/\detokenize{#1}}{\url{https://doi.org/\detokenize{#1}}}}
\usepackage{graphicx}
\usepackage{listings}
\usepackage{color}
\usepackage{amsmath}
\usepackage{multirow}
\usepackage{makecell}
\usepackage[labelfont=bf]{caption}
\usepackage[labelfont=bf]{subcaption}
\usepackage[normalem]{ulem}

\begin{document}

\title{Neural modelling of dynamic systems with time delays based on an adjusted NEAT algorithm}

\titlerunning{Neural modelling of dynamic systems with time delays}

\author{Krzysztof Laddach\inst{1}\orcidID{0000-0001-9122-2167} \and
Rafa\l{} \L{}angowski\inst{1}\orcidID{0000-0003-1150-9753}\thanks{Corresponding author - e-mail address: rafal.langowski@pg.edu.pl}}

\authorrunning{K. Laddach and R. \L{}angowski}

\institute{Department of Intelligent Control and Decision Support Systems, Digital Technologies Center, Gda\'nsk University of Technology, G. Narutowicza 11/12, 80-233 Gda\'nsk, Poland}

\maketitle

\begin{abstract}
A problem related to the development of an algorithm designed to find an architecture of artificial neural network used for black-box modelling of dynamic systems with time delays has been addressed in this paper. The proposed algorithm is based on a well-known NeuroEvolution of Augmenting Topologies (NEAT) algorithm. The NEAT algorithm has been adjusted by allowing additional connections within an artificial neural network and developing original specialised evolutionary operators. This resulted in a compromise between the size of neural network and its accuracy in capturing the response of the mathematical model under which it has been learnt. The research involved an extended validation study based on data generated from a mathematical model of an exemplary system as well as the fast processes occurring in a pressurised water nuclear reactor. The obtaining simulation results demonstrate the high effectiveness of the devised neural (black-box) models of dynamic systems with time delays.

\keywords{Neural modelling \and Neural network architecture search \and PWR black-box model.}

\end{abstract}


\section{Introduction}\label{sec:introduction}

Nowadays, an effective handling of majority industrial plants requires advanced algorithms based on the proper mathematical models of processes that occur in them. These algorithms perform different tasks such as diagnostics, monitoring, estimation, control, etc. It is known that the model makes it possible, e.g., to predict the trajectories of selected process variables, also in unacceptable situations in a real plant, to analyse the behaviour of a given process, or to consider various control strategies, etc. However, the complexity of many processes is significant, and the phenomena occurring in them are sophisticated. Therefore, delivering the system white-box model might become very difficult and even impossible. Thus, developing an alternative model such as a black-box or grey-box may be justified and reasonable \cite{Roffel2006}. These models are built based on observation of behaviour of a given system. There are different tools to create black-box models. Artificial neural networks (ANNs) are one of the most common and have produced excellent results in many fields of science \cite{Azar2015}. The neural (black-box) model is based on ANN, which according to the theorems of Kolmogorov and Cybenko has the ability to represent any mathematical function \cite{Cybenko1989,Kolmogorov1957}. However, to achieve it, an architecture of ANN (parameters and hyper-parameters) have to be selected for each unique function. This problem is known in the literature, where there are many ways to solve it, although it is still an open issue. One of the common approach to select ANN's architecture is to use neuroevolution \cite{Miikkulainen2017}, i.e. genetic or evolutionary algorithms \cite{Elsken2019,Laddach:2022,Stanley2019}. One of the most popular neuroevolutionary algorithms is NeuroEvolution of Augmenting Topologies (NEAT) \cite{Stanley2002}, which has been and still is used and modified to adapt its operation to new tasks \cite{Papavasileiou2020}. 

In this paper, the authors' neuroevolution method based on an adaptation of the NEAT algorithm to build a black-box model of a dynamic system with time delays is presented. The NEAT algorithm has been adjusted by allowing additional connections within ANN and developing original specialised evolutionary operators. An algorithm has been developed and verified by simulation that makes it possible to find ANN architecture that represents a compromise between its size and its accuracy in capturing the response of the model under which it has been learnt. The derived algorithm has been marked as dNEAT. To summarise, the main contribution of this work is to develop and verify the dNEAT algorithm that enables the selection of ANN architecture for black-box modelling of dynamic systems with time delays.

As applications, the single-input single-output (SISO) exemplary system and SISO model of the fast processes in a pressurised water reactor (PWR) are taken into account. A PWR is a non-linear, spatial and non-stationary plant whose processes are characterised by multi-scale and complex dynamics and involve delays associated with occurring delayed neutrons. Thus, there are many different mathematical models of PWR that are used depending on the aim, e.g., modelling of physical processes, diagnostics, on-line monitoring, etc. For example, seven modelling methods of processes taking place in a PWR are distinguished in \cite{Li2016}. Starting from the simplest one allows obtaining point-parameters models of low complexity, through a one-dimensional, three-dimensional, multi-point modelling based on fractional-order calculus, up to building models consisting of sub-models and neural models. However, the models built in white-box way 'pay' for their accuracy with a significant degree of complexity. On the other hand, the 'intelligent identification' of ANN-based processes in a PWR, and therefore also the building of ANN-based black-box models, is a topic enjoying considerable activity in the scientific research space \cite{Li2016}. Available in the literature neural reactor models, e.g., \cite{Boroushaki2017,Khalafi2009}, systems for identifying reactor parameters or states, e.g., \cite{Kim2017,Moshbar2014}, or neural controllers used in a PWR control structures, e.g., \cite{Coban2014} confirm the potential of ANNs in this domain. Also, the fact that the strong demand for models of these plants results from the lack of possibility to perform experiments on a real nuclear reactor. Thus, it seems that models of a PWR that would allow numerous tests to be carried out in a short time and at a low computational cost would make it possible to design control, diagnostic and safety systems in which various scenarios of events and the controls would be considered. From the research point of view addressed in the paper, a PWR model has been used only to generate learning and validation data. Hence, for this reason, a detailed description of it is not presented. 


\section{Problem statement}\label{sec:problem_statement}

\begin{figure}
\centering
    \includegraphics[width=0.7\textwidth]{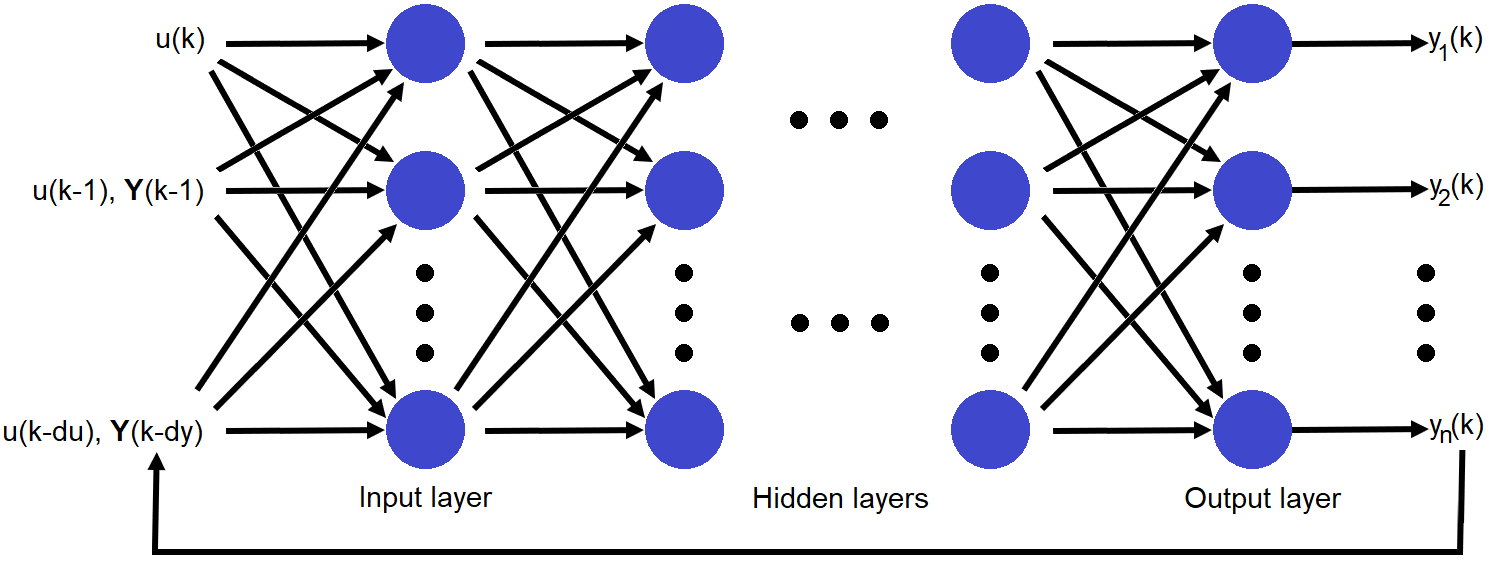}
    \caption{An architecture of exemplary RNN.} 
    \label{fig:modelSieci}
\end{figure}

As it has been mentioned, the main aim of this work is to develop and verify the authors' algorithm - the dNEAT for black-box modelling purposes of dynamic systems with time delays. As a type of ANN, a recurrent network (RNN) has been selected. In RNNs, signals may flow in both directions, i.e. from input to output and vice versa. As a result, RNNs have an internal state that depends on the current input data and the previous network states. A recurrent network has been selected because, similar to considered plants' dynamics, they have an internal feedback. An architecture of exemplary RNN is presented in Fig.~\ref{fig:modelSieci}.

Searching for a network with such architecture as in Fig.~\ref{fig:modelSieci} is equivalent to looking for a non-linear auto-regressive exogenous discrete model of the process that is based on its time-delayed input and output signals:
\begin{equation}
\begin{split}
    y(k)=f (y(k-1),y(k-2),y(k-3),...,y(k-dy),\\ 
    u(k),u(k-1),u(k-2),u(k-3),...,u(k-du)),
    \label{eq1}
\end{split}
\end{equation}
where: $y(\cdot)$, $u(\cdot)$ are the output and input at the discrete-time instant specified by $(\cdot)$, respectively; $f(\cdot)$ denotes the function specified by $(\cdot)$; $du$, $dy$ are the maximal levels of input and output (recurrence) delays, respectively; $k$ is the discrete-time instant.

The following parameters and hyper-parameters (architecture of the network) should be determined to create a black-box model using RNN: the number of neurons in assumed single hidden and output layers (so-called hidden and output neurons); the existence of connections between individual neurons; the delay levels of $du$ and $dy$; the neurons' model, i.e. type and parameters of excitation function and activation (transfer) function; the connections' model; and the values of weights of individual connections. Some of the above have been restricted to predetermined values (hyper-parameters). The excitation function is assumed, classically, as a weighted sum of inputs to a given neuron. In turn, the activation function of neurons in the hidden layer is assumed to a sigmoidal bipolar. This choice has been made because the considered RNNs are intended to represent the operation of continuous plants; thus, the activation function should be also continuous. In turn, the neuron activation function in the output layer is assumed as linear with a directional coefficient equal to one. It is since its task is to linearly transform the sum of the outputs of the previous layer, i.e. the excitation signal. The connection model has been adopted as a unit function without delays. To fully define the architecture of RNN, it is still necessary to select the parameters: the number of hidden neurons, the existence of connections between individual neurons, the values of $du$ and $dy$, and the weights of individual connections. The proposed dNEAT algorithm is used for these purposes. It is worth adding that the NEAT algorithm (dNEAT as well) provides parallel optimisation of an architecture of the neural network achieved during their evolutionary development \cite{Stanley2002}.


\section{dNEAT algorithm}\label{sec:dneat_des}

This section describes the dNEAT algorithm by pointing out the changes made relative to the NEAT algorithm \cite{Stanley2002}. Hence, it focuses only on the parts that are new concerning the NEAT algorithm. The python code of the NEAT algorithm from \cite{neat-python} has been used and errors in it have been corrected, so that it works according to the original description from \cite{Stanley2002}. The codes of the dNEAT and mentioned the NEAT are available in \cite{dneat_neat_code}. The main algorithms' parameters are:
\begin{itemize}
    \item $pop_{size} = 100$ - the target number of individuals in population;
    \item $weight_{init_{mean}} = 0$, $bias_{init_{mean}} = 0$ - the Gaussian distribution mean values for drawing initial values of weights and biases;
    \item $weight_{init_{stdev}} = 0.5$, $bias_{init_{stdev}} = 0.5$ - the Gaussian distribution standard deviations for drawing initial values of weights and biases;
    \item $du_{init-max} = 20$; $dy_{init-max} = 20$ - the upper ranges of intervals for drawing values $du$ and $dy$;
    \item $bias_{mutate_{power}} = 0.033$ - the zero-centred Gaussian distribution standard deviation for drawing value of bias mutation;
    \item $bias_{mutate_{rate}} = 0.2$, $bias_{replace_{rate}} = 0.2$ - the probability that mutation will change the bias of a node by adding or assigning a random value;
    \item $weight_{mutate_{power}} = 0.1$ - the zero-centred Gaussian distribution standard deviation for drawing value of weight mutation;
    \item $weight_{mutate_{rate}} = 0.6$, $weight_{replace_{rate}} = 0.05$ - the probability that mutation will change the weight by adding or assigning a random value;
    \item $conn_{add_{prob}} = 0.2$, $conn_{delete_{prob}} = 0.2$ - the probability that mutation will add or delete a connection between existing neurons;
    \item $enabled_{mutate_{rate}} = 0.2$ - the probability that mutation will change the enabled status of a connection;
    \item $node_{add_{prob}} = 0.2$, $node_{delete_{prob}} = 0.2$ - the probability that mutation will add a new neuron or delete an existing neuron; 
    \item $compatibility_{threshold} = 2.3$ - the individuals whose genomics distance is less than this threshold are considered to be in the same species;
    \item $max_{stagnation} = 25$ - the species, in which the best individual has not shown improvement over more than this number of generations will be considered stagnant and removed;
    \item $species_{elitism} = 3$ - number of species that will be protected from stagnation;
    \item $elitism = 10$ - the number of most-fit individuals in each species that will be preserved as-is from one generation to the next;
    \item $survival_{threshold} = 0.25$ - the fraction for each species allowed to reproduce each generation;
    \item $du_{mutate_{rate}} = 0.2$, $dy_{mutate_{rate}} = 0.2$ - the probability that mutation will change the delay levels;
    \item $du_{mutate-power} = 2$, $dy_{mutate-power} = 2$ - the specifies the intervals from which the delay mutation values are drawn.
\end{itemize}


\subsection{Initialisation} \label{subsec:init}

Each individual is initialised as the RNN consisting of one hidden and one output neuron. The number of output neurons reasons from the number of mapped trajectories. During initialisation, each input neuron is connected to each hidden neuron, and each hidden neuron is connected to each output neuron. The initial delays $du$ and $dy$ are drawn with uniform probability distribution from  ranges of $[0, du_{init-max}]$ and $[0, dy_{init-max}]$. The initial values of the weights and biases are drawn using Gaussian distributions with parameters given in section \ref{sec:dneat_des}.


\subsection{Crossover} \label{subsec:cross}

The crossover operator in respect of neurons and connections between them remained the same as in the NEAT algorithm. In turn, concerning the delays levels of $du$ and $dy$ crossover way yields:
\begin{equation}
    d_{child} = round\left(r d_{parent1} + (1-r) d_{parent2}\right),
\end{equation}
where: $d_{child}$ is the value of $du$ or $dy$ for offspring; $d_{parent1}$, $d_{parent2}$ are the values of $du$ or $dy$ for parents; $r$ is a number drawn using a uniform probability distribution from the range $[0, 1]$.


\subsection{Mutation} \label{subsec:mutation}

As for crossover, the mutation concerning neurons and connections between them remained the same as in the NEAT algorithm. The mutation of delays $du$ and $dy$ proceeds as follows. If draw decided that the mutation of the corresponding delay $d(\cdot)$ should occur, the delay change $\delta d(\cdot)$ is drawn with uniform probability distribution from the range $[-d(\cdot)_{mutate-power},$ $d(\cdot)_{mutate-power}]$. Then $\delta d(\cdot)$ is rounded to an integer value. If $\delta d(\cdot)$ is zero then it is assigned a value of -1 or 1 with equal (50\%) probability. The corresponding delay is then changed according to the formula $d(\cdot) = |d(\cdot) + \delta d(\cdot)|$. Hence, if the corresponding delay level $du$ or $dy$ is decreased, then outgoing connections from the given input are removed. On the other hand, if the delay is increased, outgoing connections from the new inputs are added to all hidden neurons in the same way as during initialisation.


\subsection{Fitness function} \label{subsec:fitness}

Since the dNEAT algorithm optimises the structure of individuals directly during its operation, the function determining the fitness of individuals does not need to depend on the RNN architecture. Therefore, the fitness function yields:
\begin{equation}\label{eq:fitness_function}
    f_i = -\frac{1000}{N} \sum_{j=1}^{J}\left(y_{ij}-t_j\right)^2,
\end{equation}
where: $f_i$ is the value of fitness function for $i$-th individual; $N$ is the number of samples during learning phase; $y_{ij}$ are successive $j$-th samples of the response of the $i$-th individual; $t_{j}$ are successive $j$-th samples of the target trajectory.

It can be noticed that \eqref{eq:fitness_function} is the mean square error (MSE) of mapping the learning trajectory through the network response. This value is multiplied by a relatively large negative scaling number to magnify the discrepancy in values of the fitness function of individuals with a similar mapping performance. It is a negative number because, by definition, the better individual should achieve a higher fitness function value.


\section{Applications} \label{sec:app}

The developed dNEAT algorithm has been simulation-verified on two applications. The first one is the SISO exemplary system, whereas the second is the SISO model of fast processes in a PWR. As a result, using the dNEAT algorithm, neural (black-box) models are created for both plants.


\subsection{Application 1} \label{subsec:app1}

In this application, the black-box model of an exemplary non-linear plant with time delays described by \eqref{akademicki_rownanie} has been created:
\begin{equation}\label{akademicki_rownanie}
    x(k) = -0.05x(k-1)+0.02x(k-5)+sin\left(\frac{x(k-10)}{10}\right)+u(k-15).
\end{equation}

The input ($u(\cdot)$) trajectory during the learning phase, i.e. in the phase of obtaining the model by the NEAT and dNEAT algorithms (red line), and the corresponding target (output - $x(\cdot)$) trajectory (blue line) are presented in Fig.~\ref{akademicki_uczace}. In turn, the trajectories used in the verification phase, i.e. the phase in which the responses of the obtained neural model to the trajectories that have been not used in the learning phase are shown in Figs. \ref{akademicki_weryfikacja1} and \ref{akademicki_weryfikacja2}. In all cases, the model responses have been normalised by dividing the plant's response by 30. All the figures presented in this paper are available in \cite{dneat_neat_code}.


\subsection{Application 2} \label{subsec:app2}

\begin{figure}
\centering
    \begin{minipage}{.49\textwidth}
        \centering
        \begin{subfigure}[b]{1.0\textwidth}
            \includegraphics[width=1.0\textwidth]{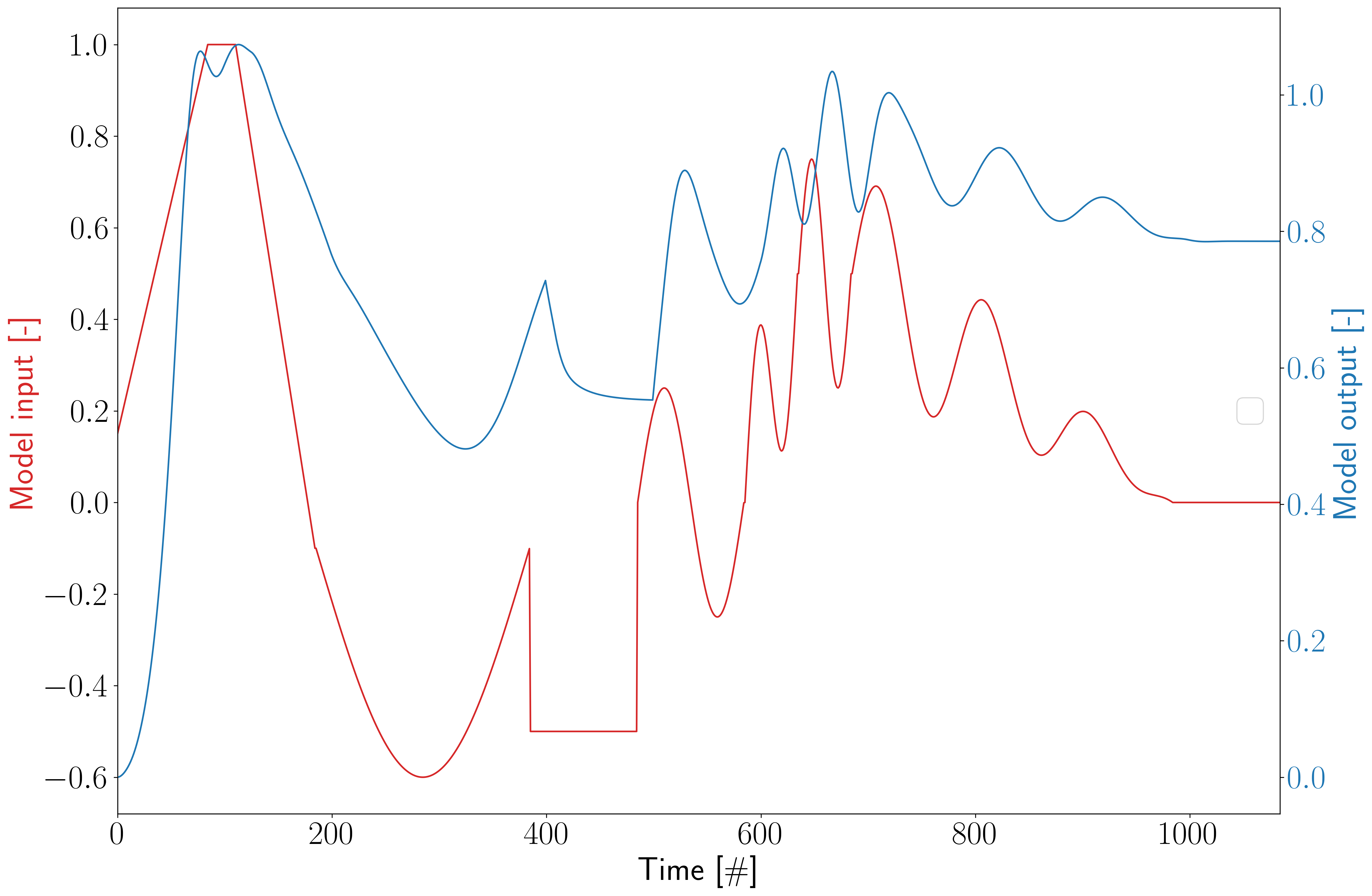}
            \caption{The learning data for application 1.}
            \label{akademicki_uczace}
        \end{subfigure}
        \begin{subfigure}[b]{1.0\textwidth}
            \includegraphics[width=1.0\textwidth]{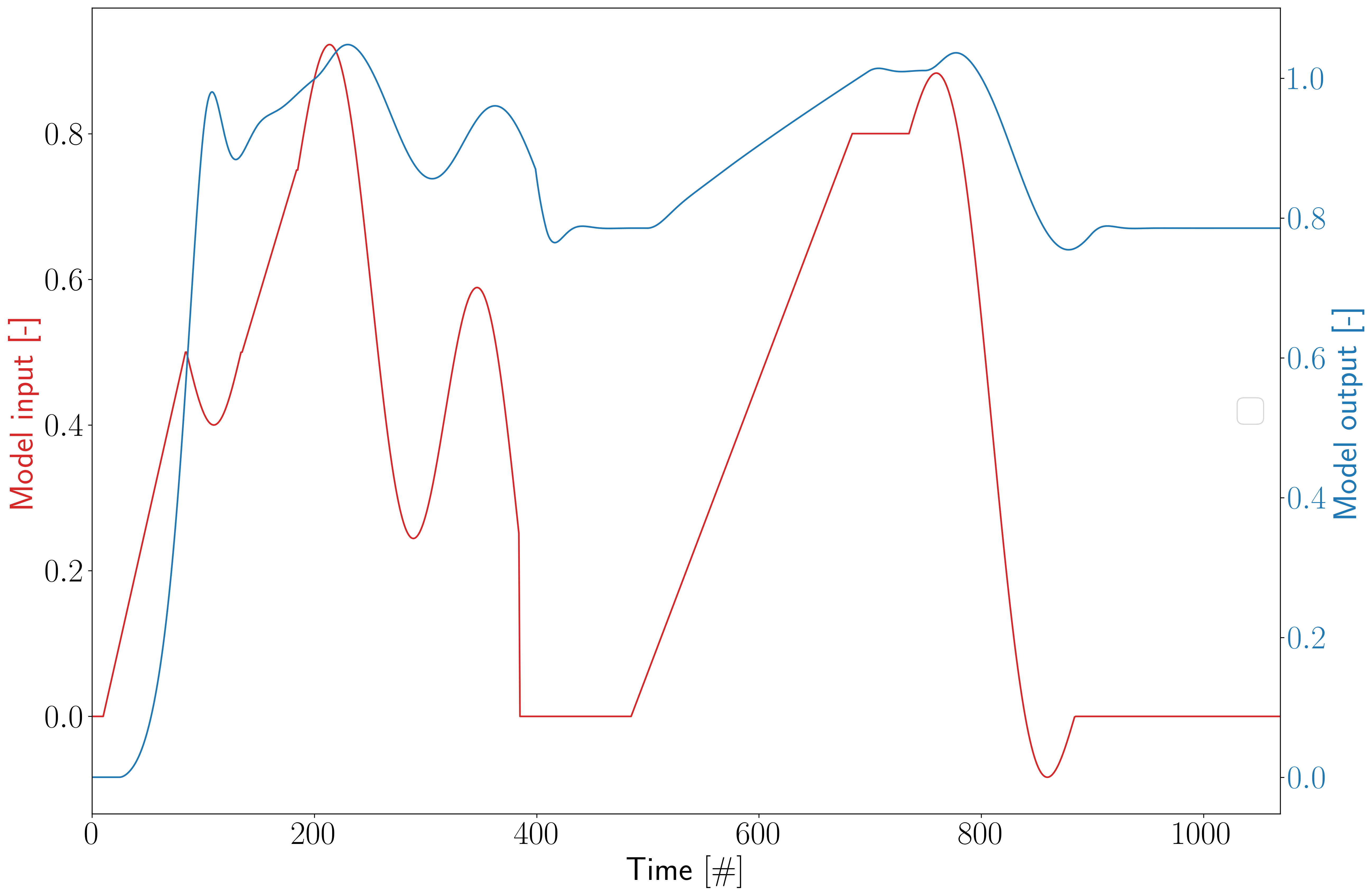} 
            \caption{The verification data for application 1 - set 1.} \label{akademicki_weryfikacja1}
        \end{subfigure}
        \begin{subfigure}[b]{1.0\textwidth}
           \includegraphics[width=1.0\textwidth]{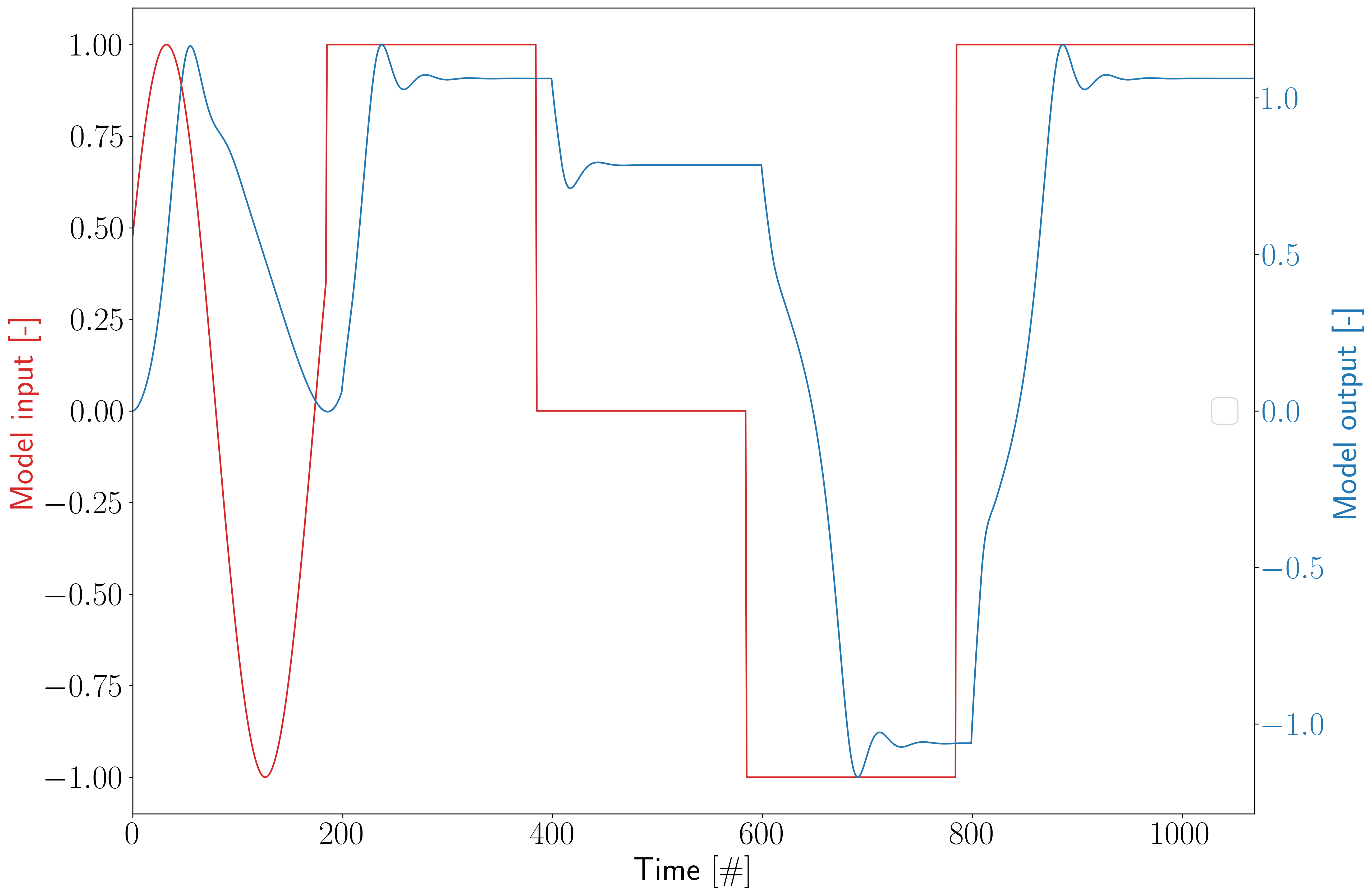} 
            \caption{The verification data for application 1 - set 2.} \label{akademicki_weryfikacja2}
        \end{subfigure}
      \caption{The learning and verification data for application 1.}
      \label{fig:data_application1}
    \end{minipage}
    \hfill\vline\hfill
    \begin{minipage}{.49\textwidth}
        \centering
        \begin{subfigure}[b]{1.0\textwidth}
            \includegraphics[width=1.0\textwidth]{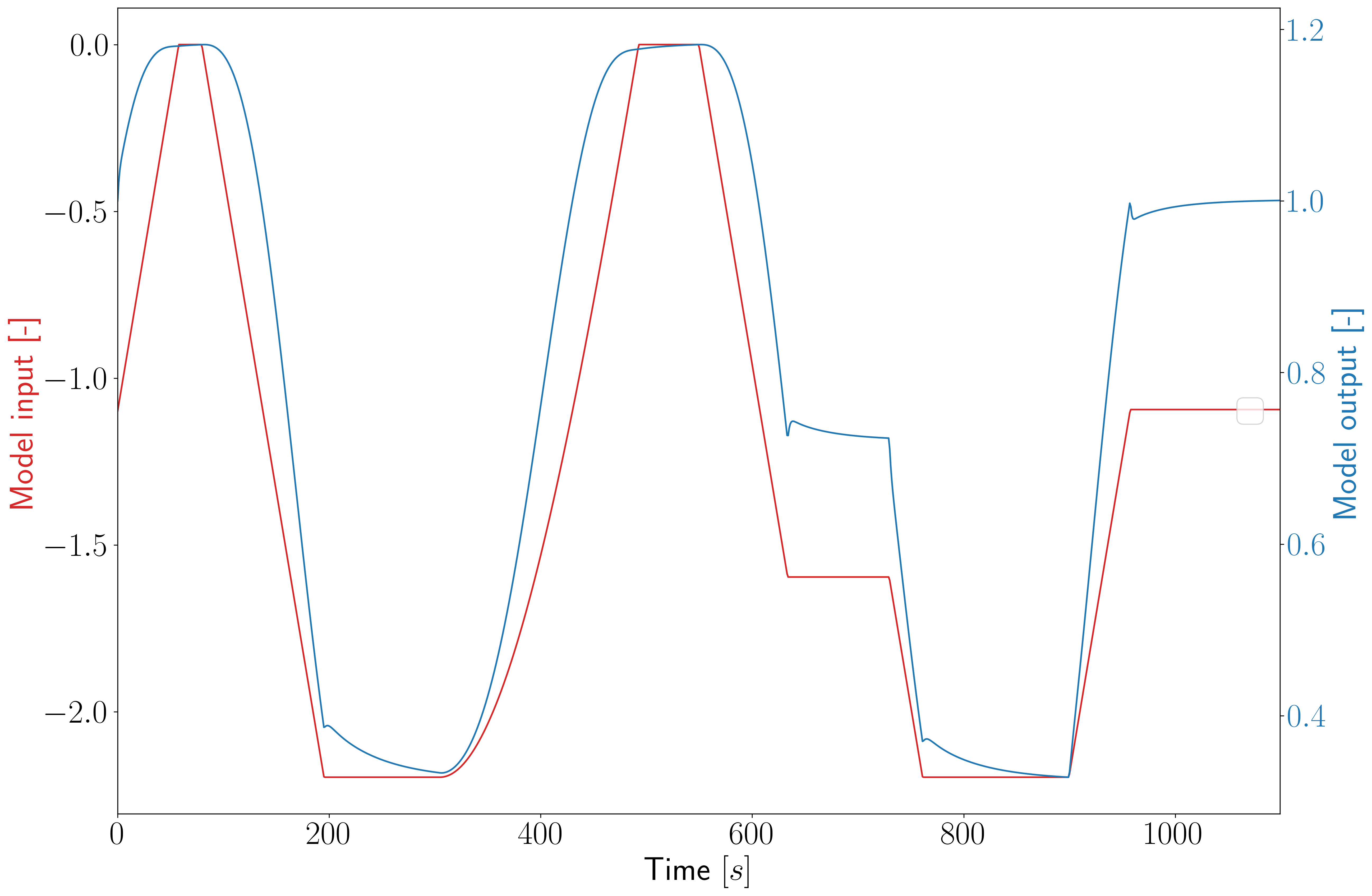}
            \caption{The learning data for application 2.}\label{pwr_uczace}
        \end{subfigure}
        \begin{subfigure}[b]{1.0\textwidth}
             \includegraphics[width=1.0\textwidth]{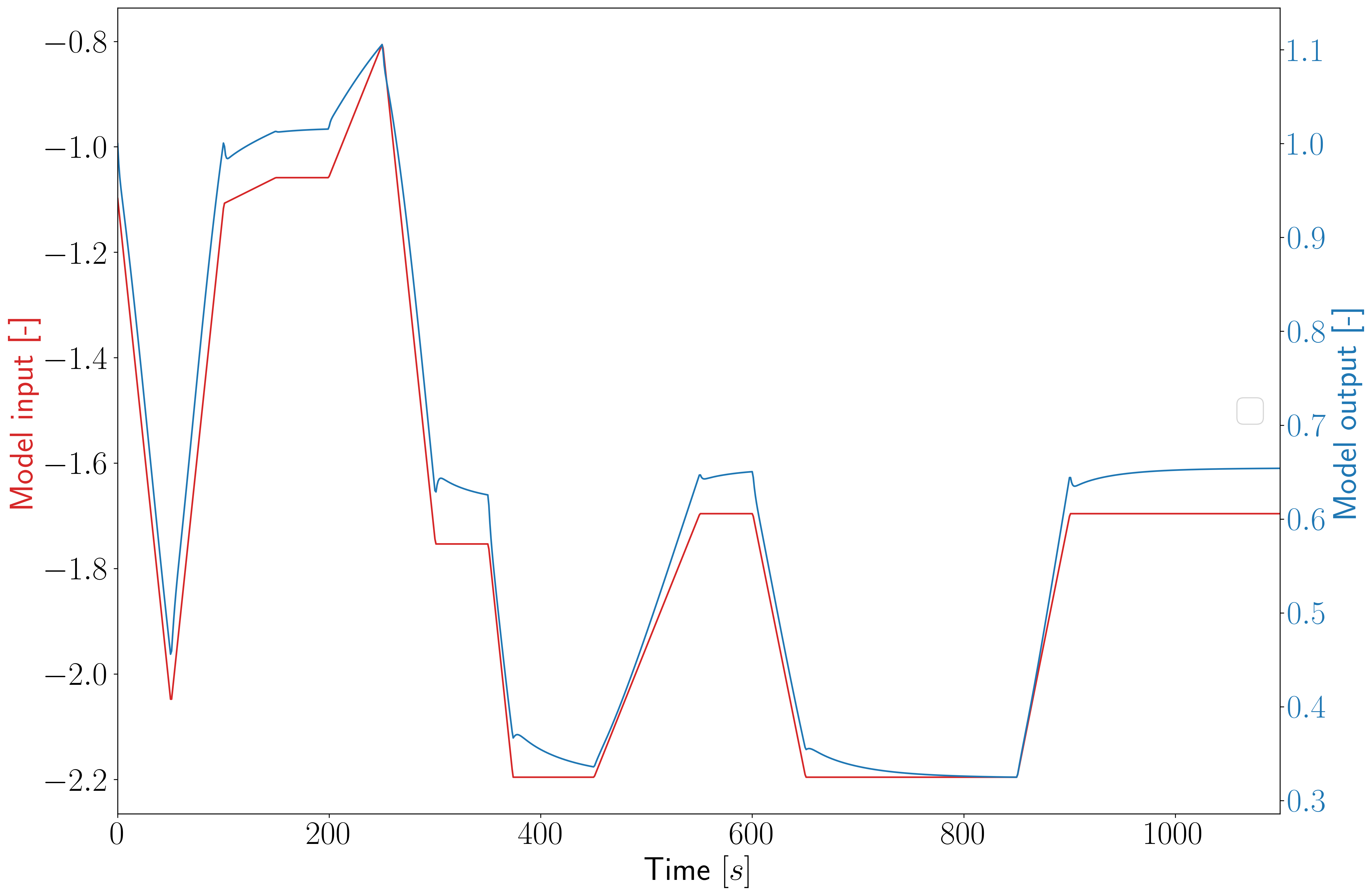} 
             \caption{The verification data for application 2 - set 1.} \label{pwr_weryfikacja1}
        \end{subfigure}
        \begin{subfigure}[b]{1.0\textwidth}
              \includegraphics[width=1.0\textwidth]{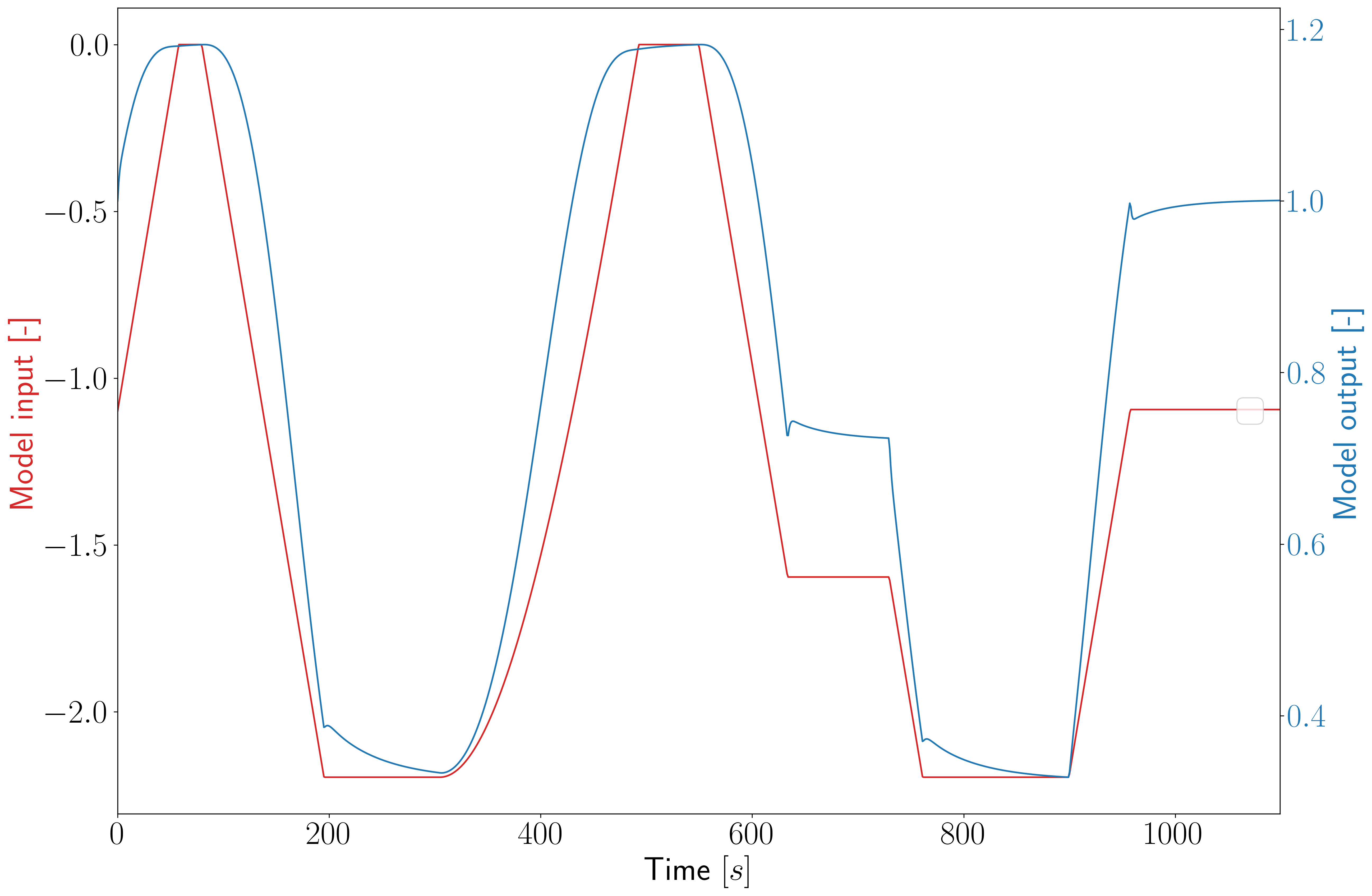} 
             \caption{The verification data for application 2 - set 2.} \label{pwr_weryfikacja2}
        \end{subfigure}
        \caption{The learning and verification data for application 2.}\label{fig:data_application2}
    \end{minipage}
\end{figure}

The second application is focused on building the SISO neural model of the fast processes in a PWR. The position of the control rods and the thermal power, scaled by dividing by the nominal power of a PWR have been selected as the input and output signals, respectively. The input trajectory during the learning phase (red line) and the corresponding output trajectory (blue line) are presented in Fig.~\ref{pwr_uczace}. In turn, the trajectories used in the verification phase are shown in Figs. \ref{pwr_weryfikacja1} and \ref{pwr_weryfikacja2}. These trajectories are taken from \cite{Laddach:2022}.


\section{Results} \label{sec:results}

\begin{figure}
  \centering
  \begin{subfigure}[b]{0.8\textwidth}
     \includegraphics[width=1.0\textwidth]{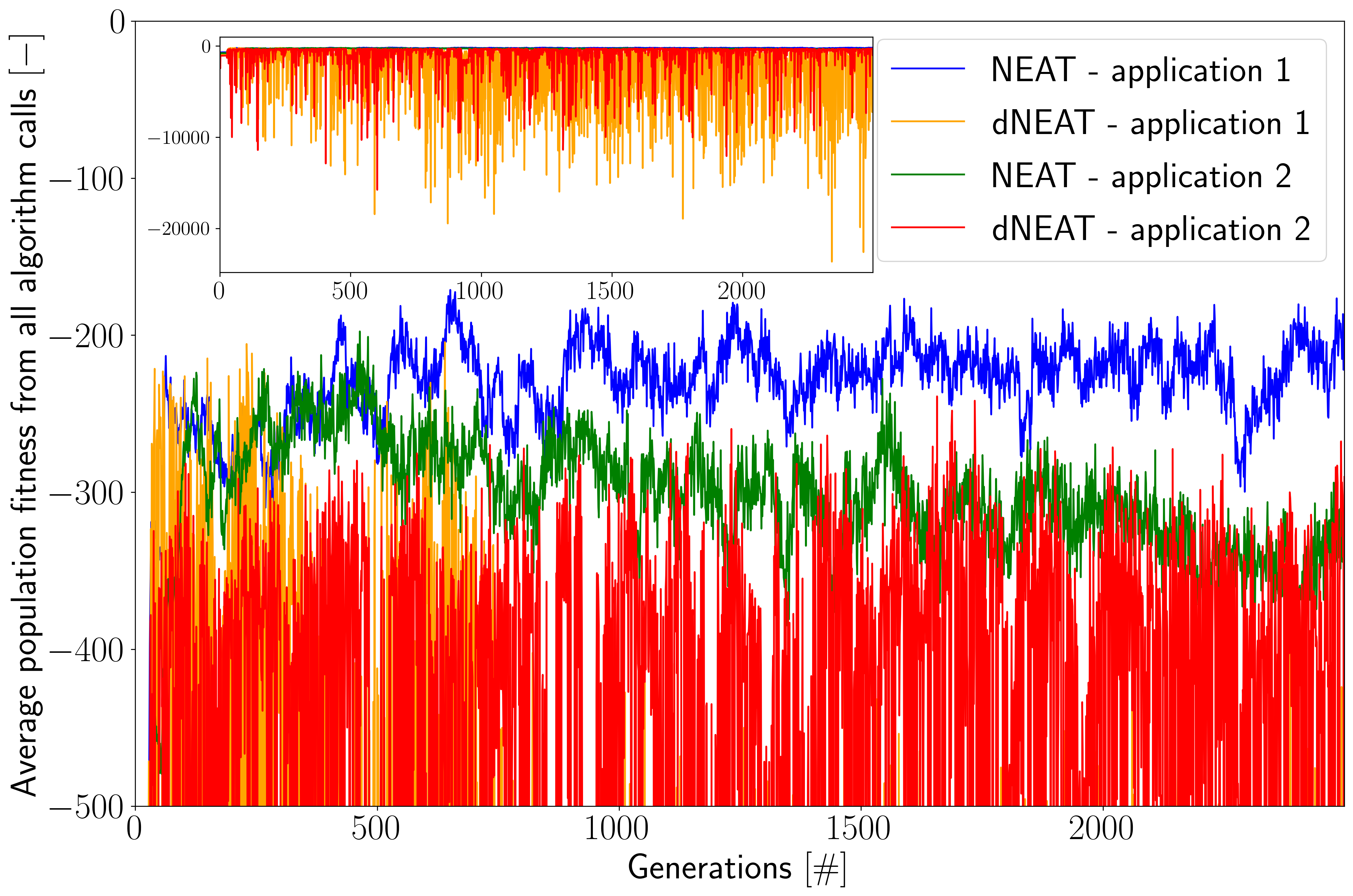}
    \caption{The average value of population fitness function.}\label{sredni_fit_generacji}
  \end{subfigure}
  \begin{subfigure}[b]{0.8\textwidth}
      \includegraphics[width=1.0\textwidth]{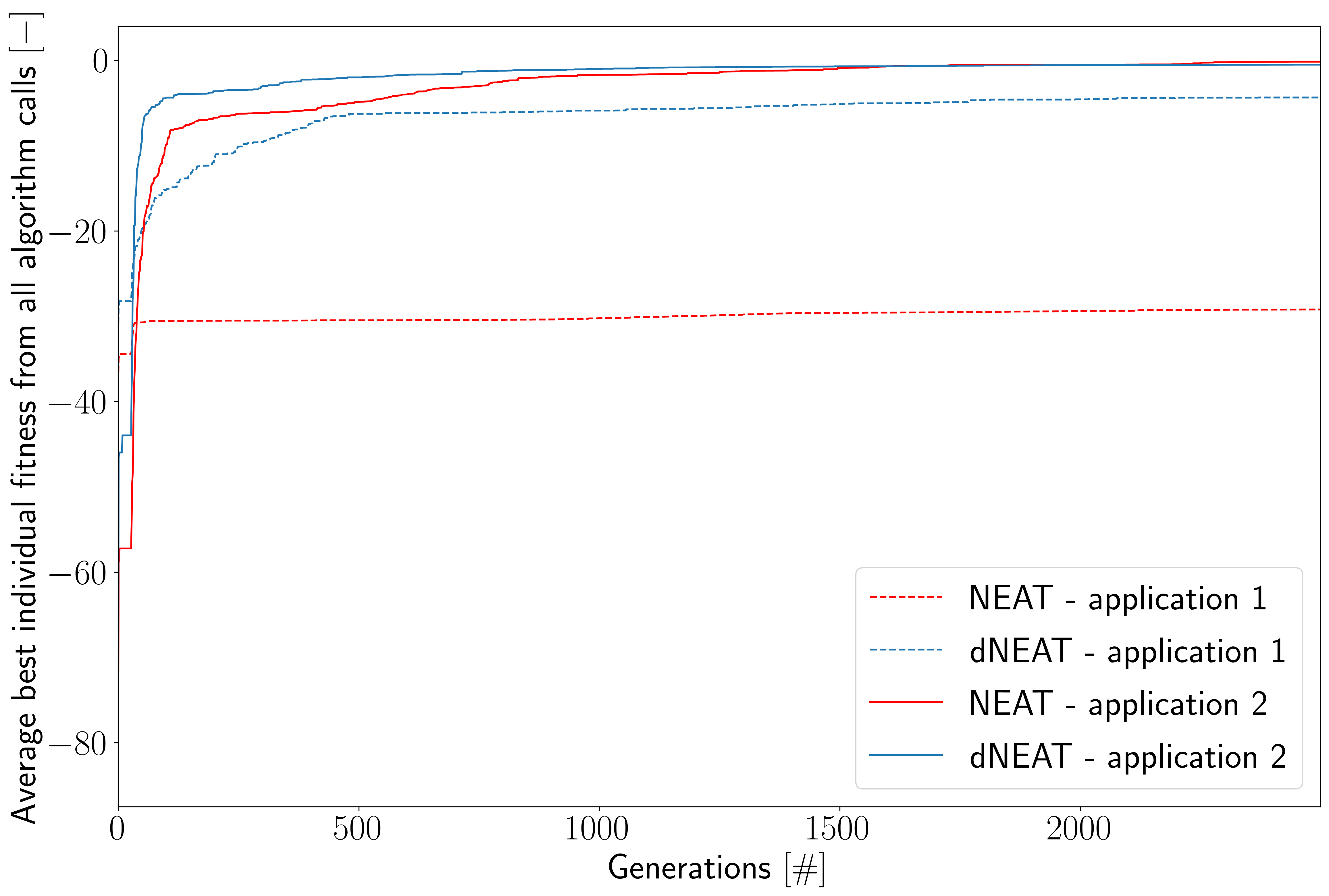} 
      \caption{The average value of fitness function of the best individuals.} \label{sredni_fit_najlepszych}
  \end{subfigure}
  \caption{The trajectories of average values of the fitness functions.}\label{fig:results1}
\end{figure}

\begin{figure}
\centering
    \begin{minipage}{.49\textwidth}
        \centering
        \begin{subfigure}[b]{1.0\textwidth}
            \includegraphics[width=1.0\textwidth]{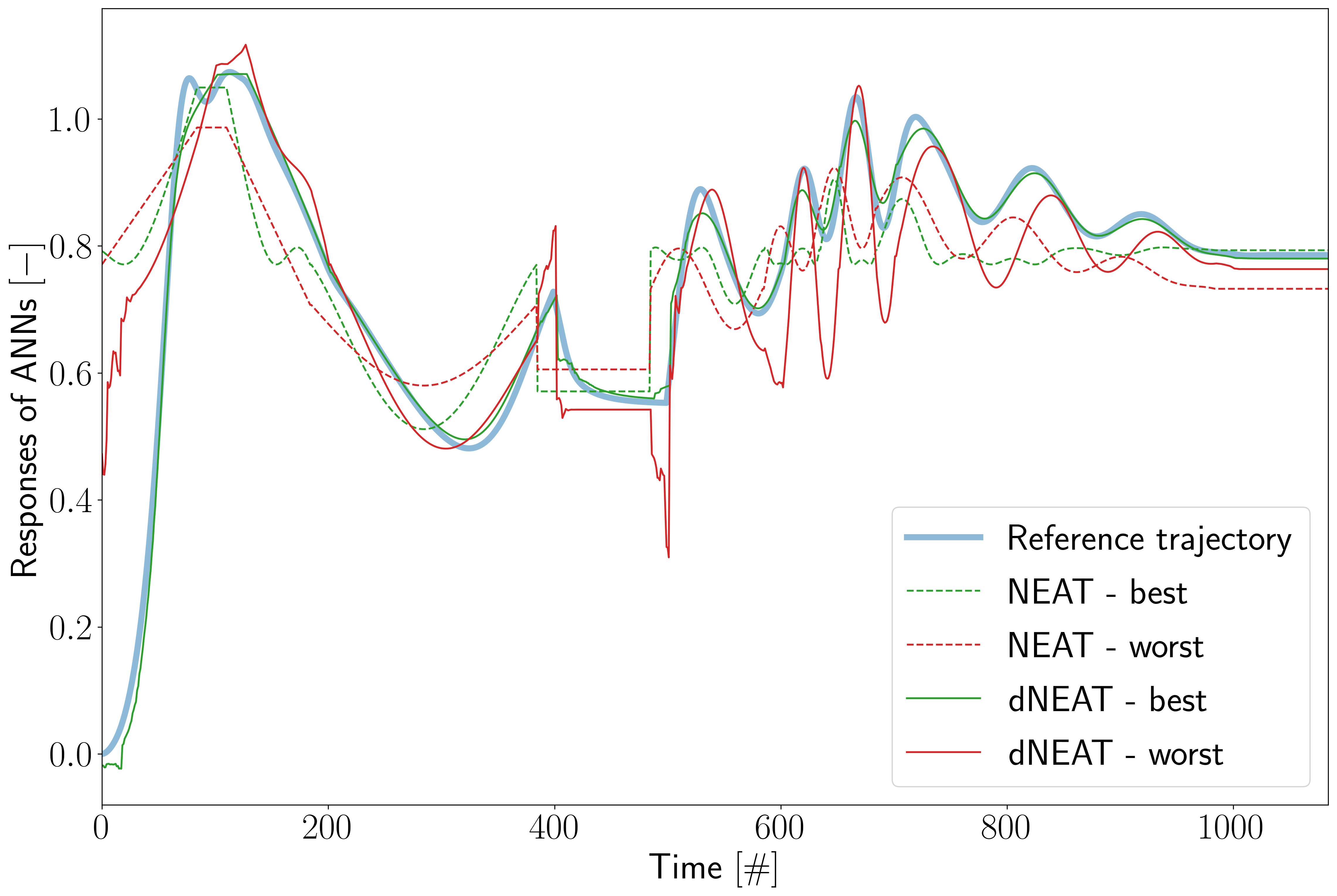}
            \caption{The responses of 'winners' in the learning phase for application 1.}
            \label{aka_app1_best_trening}
        \end{subfigure}
        \begin{subfigure}[b]{1.0\textwidth}
            \includegraphics[width=1.0\textwidth]{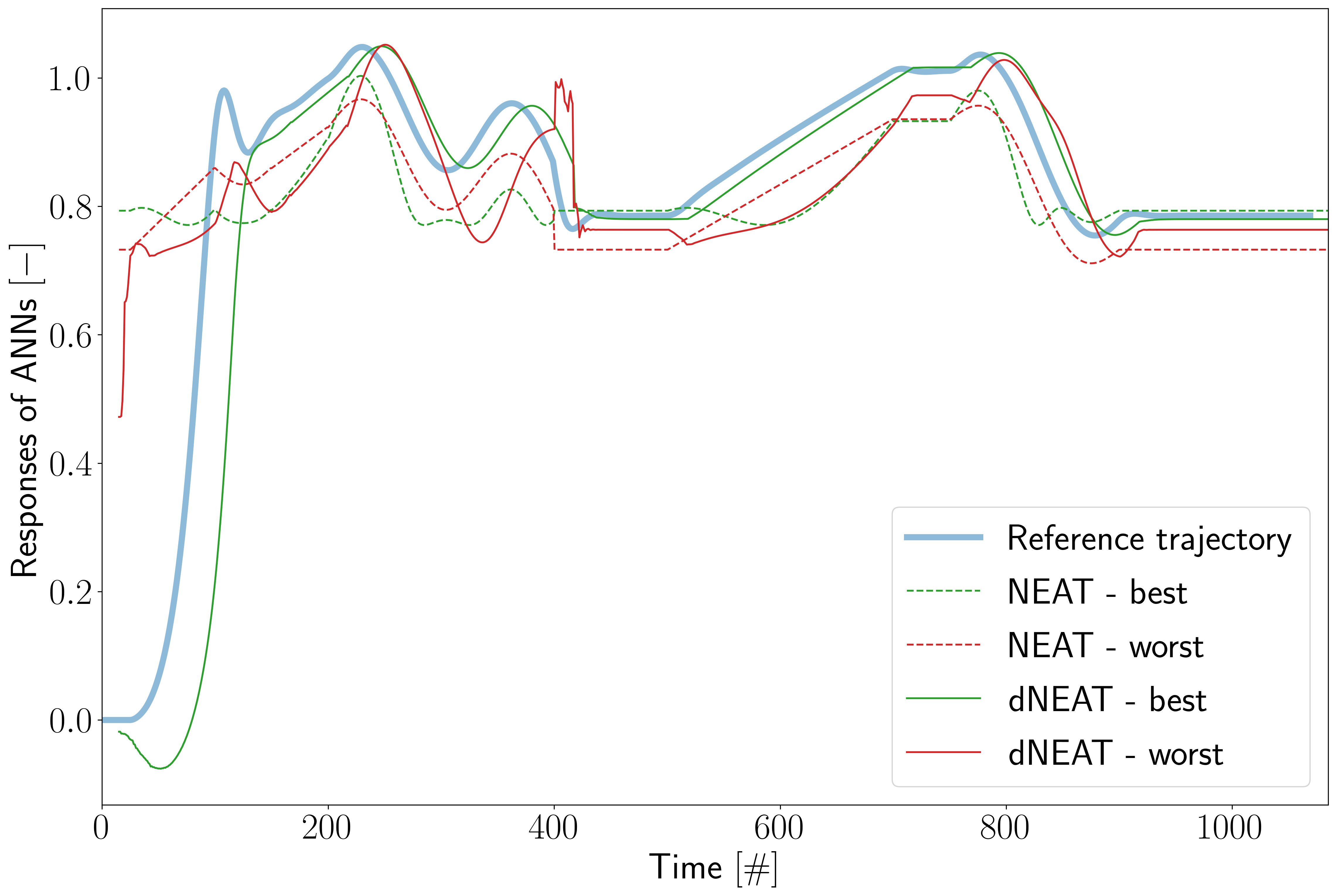} 
            \caption{The responses of 'winners' in the verification phase - set 1 for application 1.} \label{aka_app1_best_ver1}
        \end{subfigure}
        \begin{subfigure}[b]{1.0\textwidth}
            \includegraphics[width=1.0\textwidth]{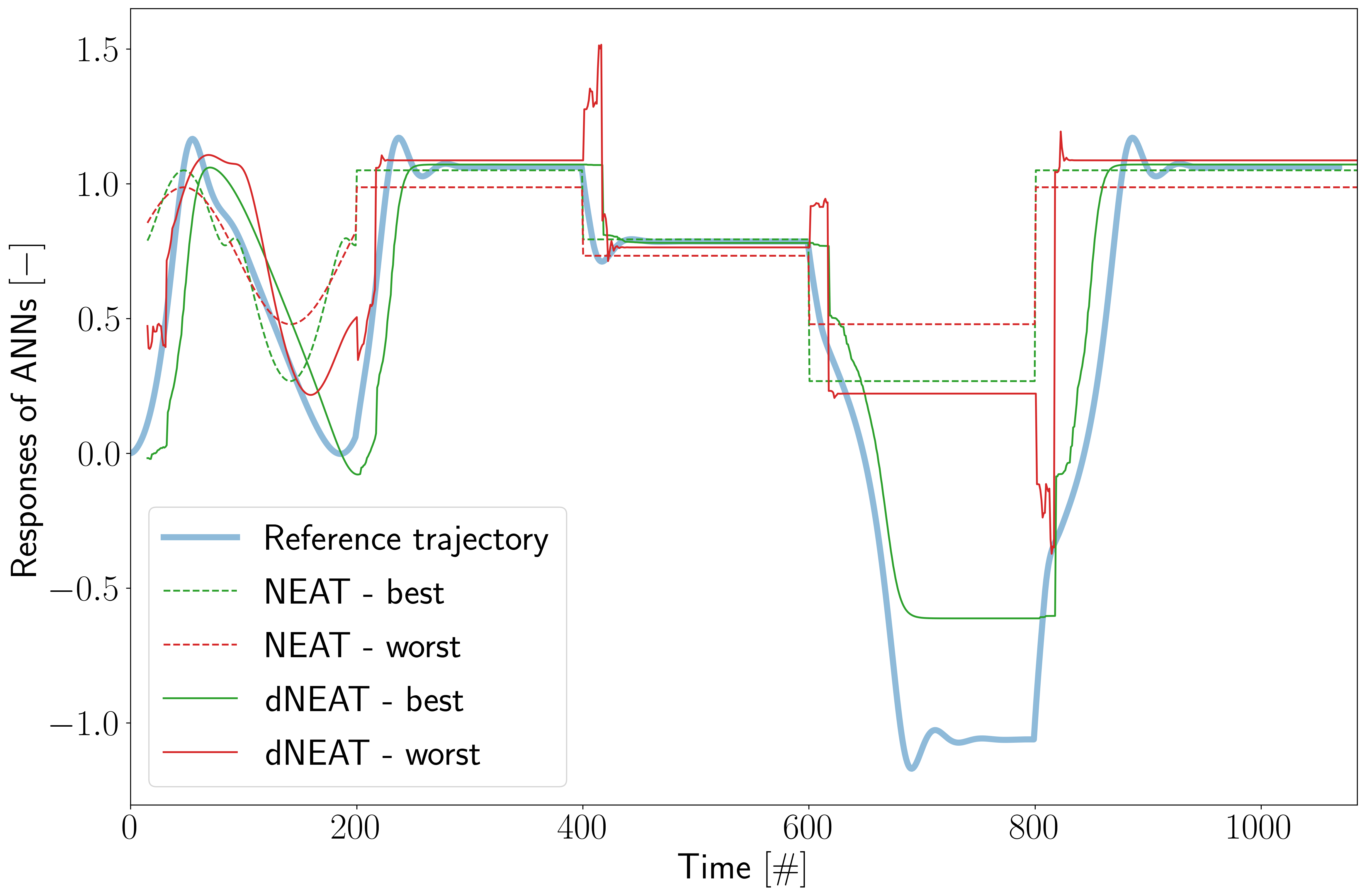} 
            \caption{The responses of 'winners' in the verification phase - set 2 for application 1.} \label{aka_app1_best_ver2}
        \end{subfigure}
        \caption{The responses of 'winners' in both phases for application 1.}
        \label{fig:data_res_app1}
    \end{minipage}
    \hfill\vline\hfill
    \begin{minipage}{.49\textwidth}
        \centering
        \begin{subfigure}[b]{1.0\textwidth}
            \includegraphics[width=1.0\textwidth]{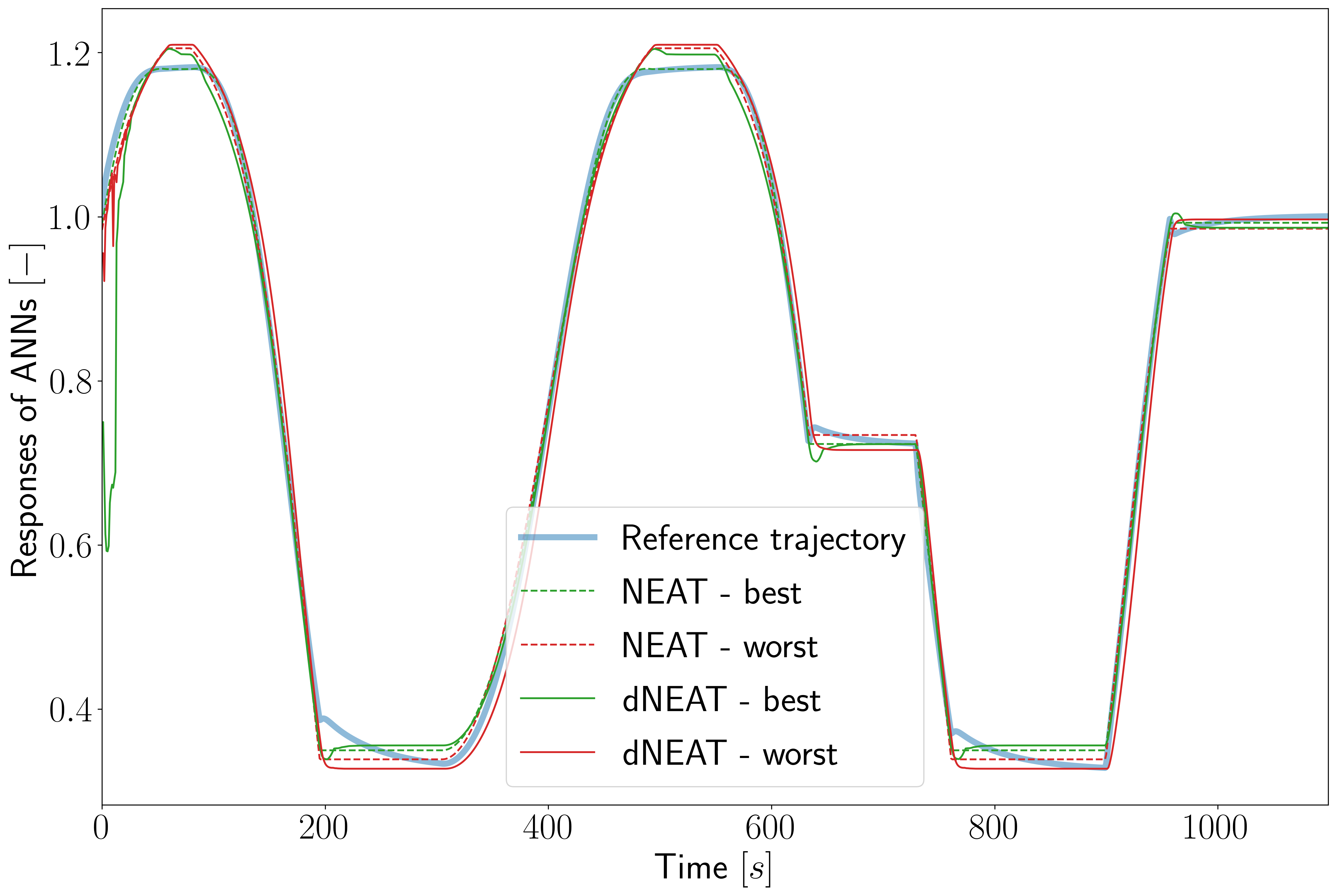}
            \caption{The responses of 'winners' in the learning phase for application 2.}
            \label{pwr_best_trening}
        \end{subfigure}
        \begin{subfigure}[b]{1.0\textwidth}
            \includegraphics[width=1.0\textwidth]{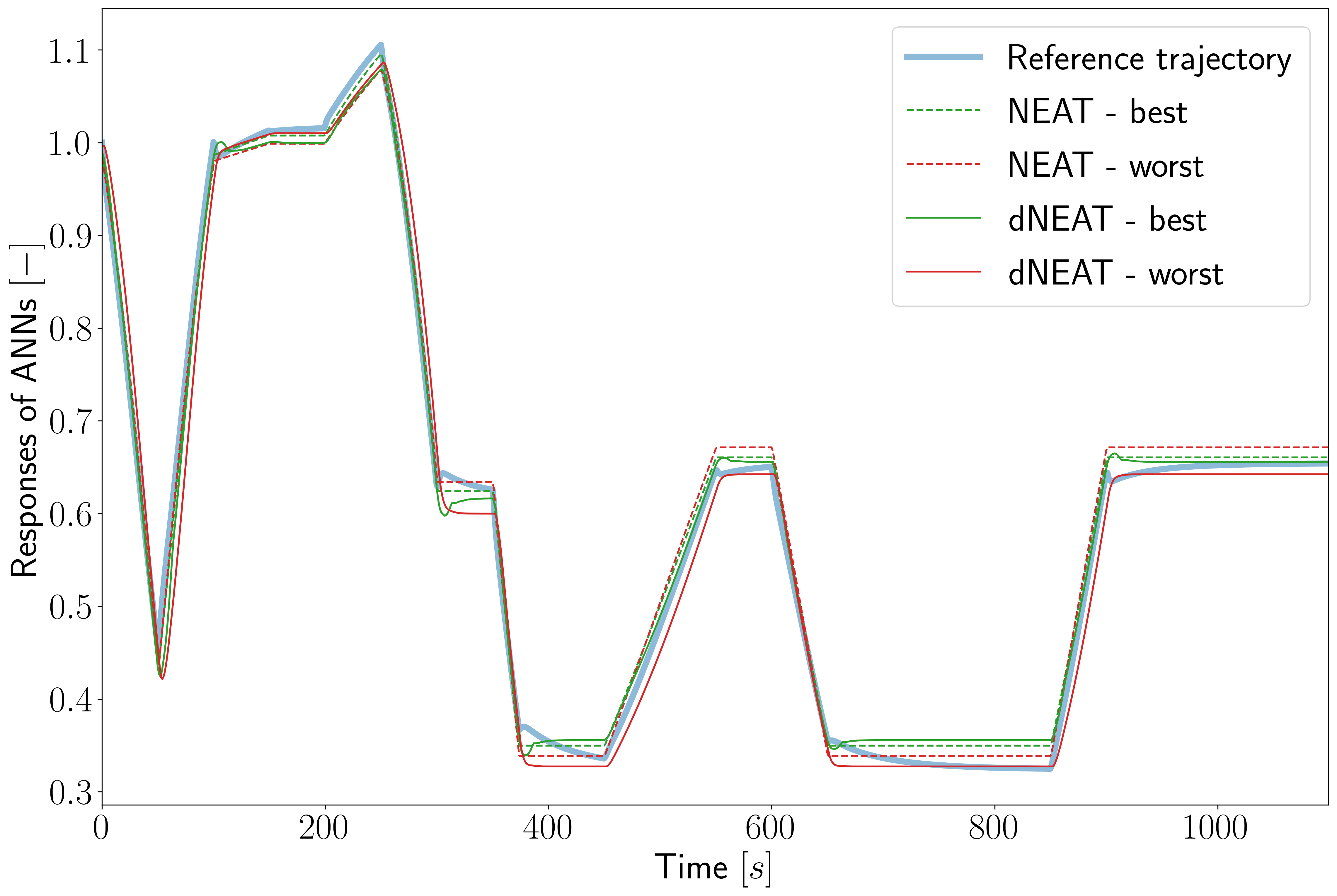} 
            \caption{The responses of 'winners' in the verification phase - set 1 for application 2.} \label{pwr_best_ver1}
        \end{subfigure}
        \begin{subfigure}[b]{1.0\textwidth}
            \includegraphics[width=1.0\textwidth]{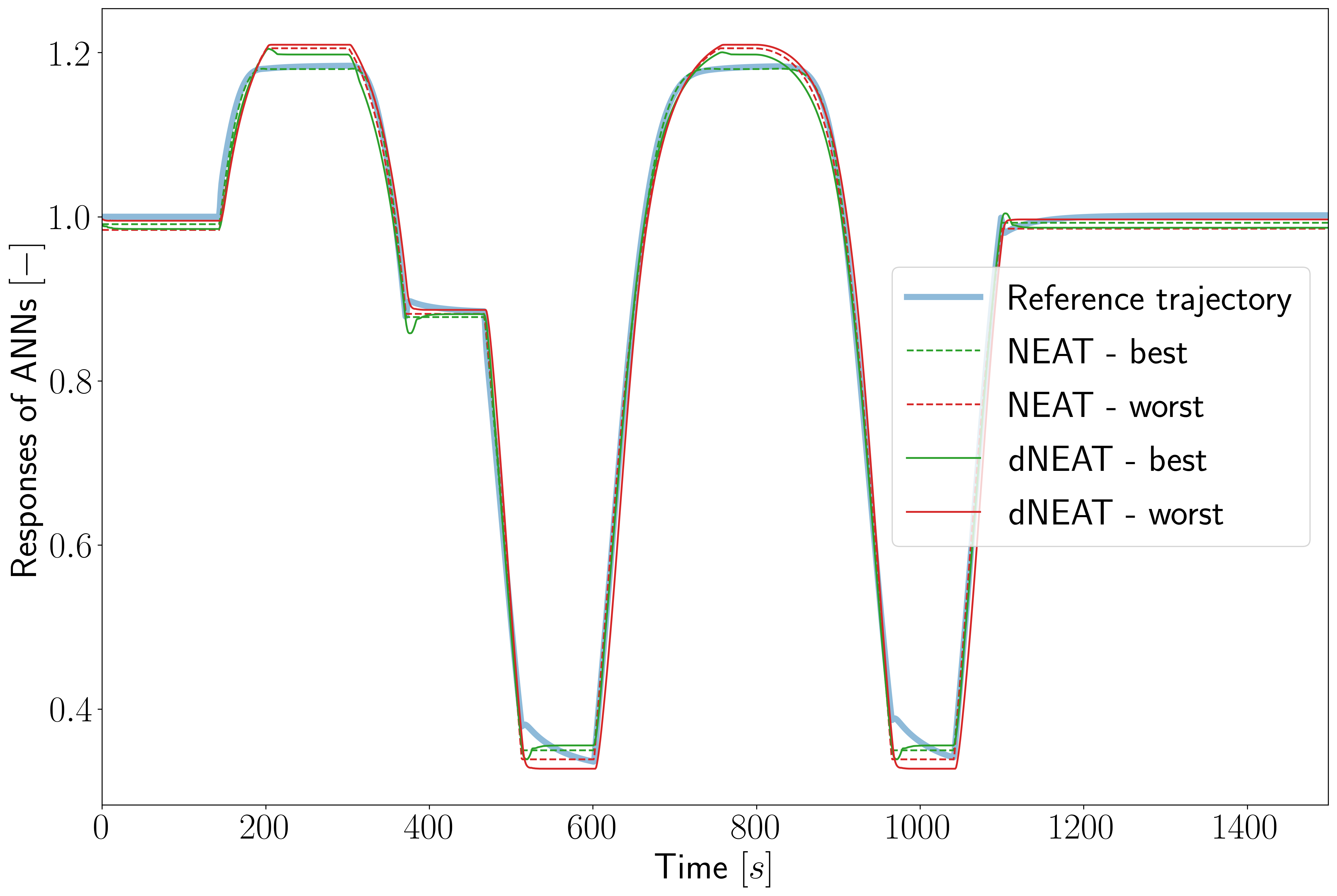}
            \caption{The responses of 'winners' in the verification phase - set 2 for application 2.} \label{pwr_best_ver2}
        \end{subfigure}
        \caption{The responses of 'winners' in both phases for application 2.}
        \label{fig:data_res_app2}
     \end{minipage}
\end{figure}

In this section, the simulation results illustrating the performance of the proposed algorithm against the results generated by the NEAT algorithm are presented. First, the learning phase is shown. Next, the verification phase is discussed. The stop condition has taken 2500 generations, and both algorithms have called 10 times. All parameters present in both algorithms have had the same values (see section \ref{sec:dneat_des}). The average value of the fitness functions of subsequent generations from all NEAT and dNEAT calls in both applications are presented in Fig. \ref{sredni_fit_generacji}. Whereas the average value of fitness functions of the best individuals in subsequent generations from all algorithms is shown in Fig. \ref{sredni_fit_najlepszych}. The responses of the neural model obtained for the learning data for the best and worst individuals by fitness function values, selected from all obtained the best individuals from each algorithm call for application 1 are shown in Fig. \ref{aka_app1_best_trening}. For simplicity, the best individuals obtained from successive algorithms calls are called 'winners'. The responses of the best and worst individuals from the winners for the verification phase are presented in Figs. \ref{aka_app1_best_ver1} and \ref{aka_app1_best_ver2}. Analogous results for application 2 are shown in Fig. \ref{fig:data_res_app2}. The MSE values of the best and worst individuals from the winners and the average MSE value of the winners from both phases for both applications are given in Table \ref{tab:wyniki_app1}.

\begin{table}
\caption{The values of MSE for applications 1 and 2.}
\label{tab:wyniki_app1}
\centering
\begin{tabular}{|c|c|c|c|c|}
    \hline
    \multirow{2}{*}{Phase} & \multirow{2}{*}{Algorithm} & \multirow{2}{*}{Description} & \multicolumn{2}{|c|}{Value $\left(\cdot 10^{-2}\right)$} \\ \cline{4-5}
    & & & Application 1 & Application 2\\
    \hline
    \multirow{6}{*}{Learning} & \multirow{3}{*}{NEAT} & the best of the winners & $2.85996$ & $0.01335$\\ \cline{3-5}
                                    & & the worst of the winners & $3.05342$ & $0.0279$\\ \cline{3-5}
                                    & & the average value of the winners & $2.92178$ & $0.01654$\\ \cline{2-5}
                                    & \multirow{3}{*}{dNEAT} & the best of the winners & $0.04096$ & $0.03113$\\ \cline{3-5}
                                    & & the worst of the winners & $1.7322$ & $0.08256$\\ \cline{3-5}
                                    & & the average value of the winners & $0.43248$ & $0.05183$\\ \cline{3-5}
    \hline
    \multirow{6}{*}{Verification - set 1} & \multirow{3}{*}{NEAT} & the best of the winners & $4.27963$ & $0.02234$\\ \cline{3-5}
                                    & & the worst of the winners & $4.54949$ & $0.0326$\\ \cline{3-5}
                                    & & the average value of the winners & $4.45215$ & $0.02278$\\ 
                                \cline{2-5}
                                    & \multirow{3}{*}{dNEAT} & the best of the winners & $0.47872$ & $0.03478$\\ \cline{3-5}
                                    & & the worst of the winners & $3.49938$ & $0.12342$\\ \cline{3-5}
                                    & & the average value of the winners & $0.98783$ & $0.08117$\\ \cline{3-5}
    \hline
    \multirow{6}{*}{Verification - set 2} & \multirow{3}{*}{NEAT} & the best of the winners & $41.99314$ & $0.00909$\\ \cline{3-5}
                                    & & the worst of the winners & $47.21822$ & $0.02924$\\ \cline{3-5}
                                    & & the average value of the winners & $42.26676$ & $0.014$\\ 
                                \cline{2-5}
                                    & \multirow{3}{*}{dNEAT} & the best of the winners & $5.58866$ & $0.02675$\\ \cline{3-5}
                                    & & the worst of the winners & $31.34926$ & $0.09197$\\ \cline{3-5}
                                    & & the average value of the winners & $16.26694$ & $0.07378$\\ \cline{3-5}
    \hline
\end{tabular}
\end{table}

Analysing the results obtained, the following conclusions can be drawn. The mean values of the fitness functions of subsequent populations indicate that for the dNEAT algorithm they have been lower than for the NEAT. However, the mean values of the fitness functions of the best individuals show that the dNEAT algorithm provides a better (or comparable) performance in finding the best individuals. Additionally, these individuals are found faster. Therefore, the dNEAT algorithm, despite searching a larger solution space and putting more emphasis on population diversity is able to find individuals better than the NEAT algorithm in the task of creating neural models of dynamic systems with time delays. The individuals found in this algorithm are more diverse through the different levels of input and output delays. Thus, the proposed solution makes it possible to find individuals acting as a black-box model of a dynamic system with time delays more efficiently concerning the NEAT algorithm.


\section{Conclusions} \label{sec:conc}

In this paper, the problem of developing an algorithm of artificial neural network architecture search for black-box modelling of dynamic systems with time delays has been investigated. In particular, the evolutionary algorithm dNEAT has been devised to build the required models. The specialised evolutionary operators and additional connections within the network ensure the proper operation of the proposed algorithm. It enabled devising the SISO neural model of the exemplary system as well as the fast processes in a PWR. The dNEAT algorithm has been implemented in the computational environment, and obtained simulation results yield satisfying performance of the produced output trajectories. Further work is needed to analyse the impact of the algorithm parameters values on its operation and to attempt to select them automatically, e.g., through machine learning.


\subsubsection{Acknowledgements} Financial support of these studies from Gda\'nsk University of Technology by the DEC-2/2020/IDUB/I.3.3 grant under the Argentum Triggering Research Grants - 'Excellence Initiative - Research University' program is gratefully acknowledged.


\bibliographystyle{splncs04}
\bibliography{bibliography}

\begin{thebibliography}{10}
\providecommand{\url}[1]{\texttt{#1}}
\providecommand{\urlprefix}{URL }
\providecommand{\doi}[1]{https://doi.org/#1}

\bibitem{Azar2015}
Azar, A.T., Vaidyanathan, S.: Computational intelligence applications in
  modeling and control. Springer, Cham, Switzerland (2015)

\bibitem{Boroushaki2017}
Boroushaki, M., Ghofrani, M.B., Lucas, C.: A new approach to spatio-temporal
  calculation of nuclear reactor cores using neural computing. Nuclear Science
  and Engineering  \textbf{155}(1),  119--130 (2017)

\bibitem{Coban2014}
Coban, R.: Power level control of the {TRIGA} {M}ark-ii research reactor using
  the multifeedback layer neural network and the particle swarm optimization.
  Annals of Nuclear Energy  \textbf{69},  260--266 (2014)

\bibitem{Cybenko1989}
Cybenko, G.: Approxmiation by superpositions of a sigmoidal function.
  Mathematics of control, signals and systems  \textbf{2},  303--314 (1989)

\bibitem{Elsken2019}
Elsken, T., Metzen, J.H., Hutter, F.: Neural architecture search. In: Hutter,
  F., Kotthof, L., Vanschoren, J. (eds.) Automated Machine Learning. The
  Springer Series on Challenges in Machine Learning, pp. 63--77. Springer,
  Cham, Switzerland (2019)

\bibitem{Khalafi2009}
Khalafi, H., Terman, M.S.: Development of a neural simulator for research
  reactor dynamics. Progress in Nuclear Energy  \textbf{51}(1),  135--140
  (2009)

\bibitem{Kim2017}
Kim, H.G., Chang, S.H., Lee, B.H.: Pressurized water reactor core parameter
  prediction using an artificial neural network. Nuclear Science and
  Engineering  \textbf{113}(1),  70--76 (2017)

\bibitem{Kolmogorov1957}
Kolmogorov, A.N.: On the representation of continuous functions of many
  variables by superposition of continuous functions of one variable and
  addition. Doklady Akademii Nauk SSSR  \textbf{114}(5),  953--956 (1957)

\bibitem{dneat_neat_code}
Laddach, K.: The developed codes of d{NEAT} and {NEAT} algorithms.
  https://git.pg.edu.pl/p969501/dneat

\bibitem{Laddach:2022}
Laddach, K., \L{}angowski, R., Rutkowski, T.A., Puchalski, B.: An automatic
  selection of optimal recurrent neural network architecture for processes
  dynamics modelling purposes. Applied Soft Computing  \textbf{116},  paper no.
  108375 (2022)

\bibitem{Li2016}
Li, G., Wang, X., Liang, B.and~Li, X., Zhang, B., Zou, Y.: Modeling and control
  of nuclear reactor cores for electricity generation: A review of advanced
  technologies. Renewable and Sustainable Energy Reviews  \textbf{60},
  116--128 (2016)

\bibitem{neat-python}
McIntyre, A., Kallada, M., Miguel, C.G., da~Silva, C.F.: Neat-python.
  https://github.com/CodeReclaimers/neat-python, accessed: 2022-02-16

\bibitem{Miikkulainen2017}
Miikkulainen, R.: Neuroevolution. In: Sammut, C., Webb, G.I. (eds.)
  Encyclopedia of Machine Learning. Springer, Boston, USA (2011)

\bibitem{Moshbar2014}
Moshkbar-Bakhshayesh, K., Ghofrani, M.B.: Development of an efficient
  identifier for nuclear power plant transients based on latest advances of
  error back-propagation learning algorithm. IEEE Transactions on Nuclear
  Science  \textbf{61}(1),  602--610 (2014)

\bibitem{Papavasileiou2020}
Papavasileiou, E., Cornelis, J., Jansen, B.: A systematic literature review of
  the successors of 'neuroevolution of augmenting topologies'. Evolutionary
  Computation  \textbf{29}(1),  7--73 (2020)

\bibitem{Roffel2006}
Roffel, B., Betlem, B.: Process dynamic and control. {M}odelling for control
  and prediction. John Wiley {\&} Sons, Inc., Chichester, West Sussex, UK
  (2006)

\bibitem{Stanley2019}
Stanley, K.O., Clune, J., Lehman, J.and~Miikkulainen, R.: Designing neural
  networks through neuroevolution. Nature Machine Intelligence  \textbf{1},
  24--35 (2019)

\bibitem{Stanley2002}
Stanley, K.O., Miikkulainen, R.: Evolving neural networks through augmenting
  topologies. Evolutionary Computation  \textbf{10}(2),  99--127 (2002)

\end{thebibliography}

\end{document}